%% file: main.tex
\documentclass[letterpaper]{article} 
\usepackage{aaai2027}  
\usepackage[hyphens]{url}  
\usepackage{graphicx} 
\usepackage{natbib}  
\usepackage{caption} 
\usepackage{algorithm}
\usepackage{algorithmic}

\usepackage{newfloat}
\usepackage{listings}
\usepackage{amsopn}
\usepackage{amsmath}
\usepackage[table]{xcolor}
\usepackage{mdframed}
\usepackage{tikz}
\usetikzlibrary{shapes.geometric, arrows.meta, positioning, calc, backgrounds, fit}

\DeclareCaptionStyle{ruled}{labelfont=normalfont,labelsep=colon,strut=off} 
\floatstyle{ruled}
\newfloat{listing}{tb}{lst}{}
\floatname{listing}{Listing}

\usepackage{booktabs}

\title{AAAI Press Anonymous Submission\\Instructions for Authors Using \LaTeX{}}
\author{
    Written by AAAI Press Staff\textsuperscript{\rm 1}\thanks{With help from the AAAI Publications Committee.}\\
    AAAI Style Contributions by Peter Patel Schneider,
    Sunil Issar,\\
    J. Scott Penberthy,
    George Ferguson,
    Hans Guesgen,
    Francisco Cruz\equalcontrib\corresponding,
    Marc Pujol-Gonzalez\equalcontrib\corresponding
}
\affiliations{
    \textsuperscript{\rm 1}Association for the Advancement of Artificial Intelligence\\

    1101 Pennsylvania Ave, NW Suite 300\\
    Washington, DC 20004 USA\\
    proceedings-questions@aaai.org
}

\title{A Neurosymbolic Approach for Constructing Planning Domain Models from Clinical Narratives}
\author {
    Ranveer Singh\textsuperscript{\rm 1}\equalcontrib\corresponding,
    Saurabh Mathur\textsuperscript{\rm 2}\equalcontrib,
    Michael Skinner\textsuperscript{\rm 1},\\
    Prasad Tadepalli\textsuperscript{\rm 3},
    Kristian Kersting\textsuperscript{\rm 2,4,5},
    Sriraam Natarajan\textsuperscript{\rm 1}
}
\affiliations {
    \textsuperscript{\rm 1}The University of Texas at Dallas, Richardson, TX, USA\\
    \textsuperscript{\rm 2}Technical University of Darmstadt, Darmstadt, Germany\\
    \textsuperscript{\rm 3}Oregon State University, Corvallis, OR, USA\\
    \textsuperscript{\rm 4}Hessian Center for Artificial Intelligence (hessian.ai), Darmstadt, Germany\\
    \textsuperscript{\rm 5}German Research Center for AI (DFKI)\\
}

\nocopyright 
\begin{document}

\maketitle

\begin{abstract}
Surgical procedures such as laparoscopic appendectomy are complex, high-stakes processes, yet formalizing their workflows for decision support remains a significant challenge. Inducing probabilistic planning domain models in this setting is particularly difficult due to the lack of structured event data and the prevalence of implicit actions in clinical narratives, which neither empirical symbolic methods nor Large Language Models (LLMs) can adequately address on their own. We introduce NSPIN, a neurosymbolic framework for inducing probabilistic planning domain models from unstructured clinical narratives. Our method extracts and imputes structured event sequences from raw text using a pretrained LLM, then induces a PPDDL model and refines its preconditions with LLM-proposed revisions, guided by empirical validation. We evaluate the approach on 2,660 laparoscopic appendectomy notes written by 9 surgeons. NSPIN yields models that generalize to unseen notes, and expert clinical review indicates its induced knowledge is largely consistent with surgical practice.  
\end{abstract}


\section{Introduction}


Surgical notes describe clinical findings and how a surgeon handled them in a specific patient, such as a perforated appendix or an unexpected abscess~\cite{mathioudakis2016keep}. However, this knowledge remains confined to the note in which it was recorded and is never used beyond that patient's chart. Learning a model of surgical actions and their preconditions and effects generalizes this experience across thousands of individual cases of a given procedure, with direct consequences for patient safety, surgical training, and equitable access to high-quality care. Such a model can underpin AI-in-the-loop systems~\cite{natarajan2025human} that help improve simulators for training residents by capturing rare complications, enable automated review of operative notes, and inform quality improvement by tracking systematic differences among surgeons and institutions. With further advances, such models could drive the development of autonomous surgical systems.

Formalizing surgical procedures requires a representation that is both expressive enough to capture their structure and interpretable enough for clinicians to validate. Formal planning languages, such as the Planning Domain Definition Language (PDDL)~\cite{ghallab1998pddl}, offer a powerful framework for this purpose, representing procedures as a set of state predicates and parameterized action schemas. This declarative symbolic logical representation is highly interpretable, allowing clinicians to validate it against established medical standards. 
\input{Figures/framework_1}

However, the transition from unstructured clinical narratives to formal symbolic logic represents a significant challenge. Extracting a formal representation from clinical notes requires handling the linguistic diversity and varied writing styles of different clinicians~\cite{hier2024efficient}. It also requires domain knowledge, since surgeons often omit information that is obvious to them but essential for maintaining a logically consistent state in a formal representation. 


We address this formalization challenge through the specific case of laparoscopic appendectomy~\cite{korndorffer2010sages} in pediatric patients. The removal of the appendix to manage acute appendicitis is among the most common operative procedures in pediatric surgery. In a large children’s hospital, pediatric surgeons may perform ten or more such procedures in a single day. This high frequency and relatively straightforward procedural workflow make the laparoscopic appendectomy an ideal domain for investigation. It provides {\em a dense, large-scale dataset of operative notes that are structurally similar enough to allow for comparison, yet linguistically diverse enough to test the limits of automated compilation}.
Using this dataset, we build a model of action preconditions and effects that generalizes across different instances of the same procedure. Beyond serving as an interpretable representation of the procedure itself, this model can also synthesize plans and narratives for downstream uses such as simulator scenario generation and training data augmentation, both of which are important building blocks for deployment-ready clinical AI systems.


We make the following key technical contributions toward interpretable, generalizable models of surgical procedures that can support trustworthy, deployable clinical AI:
(1) We introduce NSPIN, a Neurosymbolic framework that transforms unstructured surgical notes into formal, probabilistic PDDL planning domains by combining the linguistic flexibility of LLMs with the logical rigor of symbolic induction. (2) We exploit the internal world models of pretrained LLMs to fill the gaps in surgical notes, inferring the implicit clinical actions and preconditions essential for logical plan consistency. (3) We evaluate NSPIN on a database of 2,660 real-world surgical notes by 9 surgeons. Our results demonstrate that this neurosymbolic approach produces planning models that generalize to unseen surgical workflows more effectively than purely LLM-based baselines.

After introducing the necessary technical background, we present our problem and its solution. We then outline our evaluation before discussing problems for future research. 

\section{Background}
NSPIN is related to three distinct research areas: automated planning, natural language processing, and neurosymbolic learning. We observe that this type of combination of multiple AI models and systems is necessary when deploying models in high-stakes domains such as healthcare. 


\subsection{Planning Domain Models} Domain models are used to represent planning domains formally. In this work, we use Probabilistic Planning Domain Definition Language (PPDDL)~\cite{younes2004ppddl1} to represent clinical planning domains. We first introduce classical PDDL~\cite{ghallab1998pddl,ghallab2004automated}, which represents deterministic domains, and then discuss PPDDL, which extends it to represent stochasticity.

Classical PDDL defines planning problem as a tuple $\langle \mathcal{P} ,s_0, G \rangle$ where $\mathcal{P} = \langle\mathcal{F}, O, \mathcal{A}\rangle$ is the planning domain. Here, $O$ is a finite set of objects, $\mathcal{F}$ is a finite set of first-order logic predicate symbols describing properties of objects or the relations between them, $s_0$ is the initial state, $G$ is the goal condition, and $\mathcal{A}$ is a set of parameterized action schemas. Each state $s$ is a subset of the set of all possible grounded predicates $\mathcal{F}_g$~\cite{fikes1971strips}. Each action schema $a \in \mathcal{A}$ is defined as the tuple $\langle \text{pre}(a), \text{add}(a), \text{del}(a)\rangle,$ representing the conditions that must hold for the action to be applicable, the set of facts that become true after executing the action, and the set of facts that become false after executing the action respectively. Concretely, a grounded action $a_g$ is applicable in a state $s$ if its preconditions are met $\text{pre}(a_g)\subseteq s.$ Given problems formalized in this notation, classical planning aims to find a plan, defined as a finite sequence of deterministic actions $\pi = \langle a^{(1)},\dots,a^{(n)}\rangle$ that sequentially transforms the initial state $s_0$ into one satisfying the goal test $G.$

In probabilistic domains, classical PDDL must be extended by probabilistic PDDL to model domains in which the outcomes of each action are stochastic. Instead of an action $a$ having a single, guaranteed outcome defined by one set of add and delete effects, its effects are defined as a probability distribution over multiple possible outcomes. These stochastic effects are formalized as a set of tuples $\{ (p_1, \text{add}_1(a), \text{del}_1(a)), \dots, (p_k, \text{add}_k(a), \text{del}_k(a)) \}$ where $p_i$ is the probability of $i-$th outcome occurring on executing the action $a,$ and $\sum_{i=1}^k p_i = 1.$

The acquisition of these (P)PDDL models has traditionally relied on intensive manual engineering by domain experts, which is often brittle and unscalable. To automate this process, several symbolic inductive learning methods ~\cite{SAM_,le2024learning,pasula2007learning} have been proposed to extract action schemas from observed execution. These methods typically assume access to structured execution trajectories, often represented as state-action-state transitions $(s,a,s')$, from which probabilistic action dynamics can be estimated.

However, clinical narratives are rarely available in such a structured form. Converting these unstructured text documents to structured traces requires either manual annotation or automatic parsing. Moreover, human-written notes often omit critical but implicit intermediary steps~\cite{doppa2011learning}. As a result, even a perfect parser would produce incomplete and logically unsound traces. These latent dependencies and data sparsity make data-driven clinical planning domain construction challenging. Therefore, existing action-model induction methods cannot be used in this setting, since they assume structured symbolic traces or observed action sequences rather than raw clinical narratives.

\begin{figure*}[t!]
    \centering
    \includegraphics[width=\linewidth]{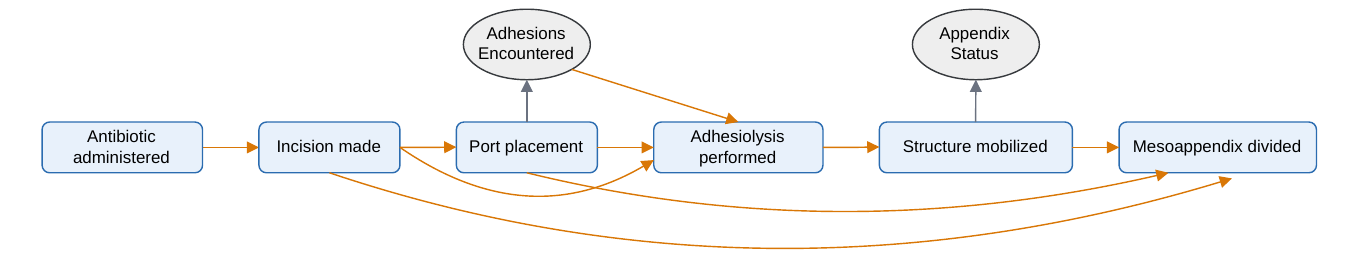}
    \caption{
   {\bf {A partial surgical sequence sampled using an action-transition model and NSPIN.}} NSPIN learns action preconditions that constrain the set of possible next actions, along with observational effects each action probabilistically produces. These emitted observations provide context for a simple data-driven action-transition model used for sampling candidate actions. \colorbox{cyan!10}{Blue} nodes represent sampled actions, \colorbox{gray!15}{gray} nodes represent emitted observations, \colorbox{orange!15}{orange} arrows indicate candidate precondition relationships, and \colorbox{gray!15}{gray} arrows indicate observation flow. In this example, port placement is followed by the observation of adhesions, which then provides contributing context for performing an adhesiolysis.
    }
    \label{fig:nspin_sampled_sequence}
\end{figure*}
\subsection{Neurosymbolic AI}
Our framework, NSPIN, takes a neurosymbolic approach~\cite{nesy} to the problem of formalizing clinical narratives into planning domains. It exploits the representation learning of neural networks with the structured reasoning of symbolic logic. Neural networks excel at extracting structured information from unstructured text~\cite{yang2022survey}. Neural models have also been used as learnable heuristics to efficiently navigate large symbolic search spaces~\cite{NN_for_planning, mangannavar2024graph}. However, the use of these methods in medicine is limited by the paucity of training data and the high cost of annotation~\cite{spasic2020clinical}.


\subsection{Pretrained LLMs}
Large language models~\cite{zhao2023survey} are a class of neural language models characterized by their large size. These models are pretrained on large natural language corpora, allowing them to process text and handle linguistic diversity without requiring additional training. As a result, these models have been used as zero-shot information extractors to handle non-standard shorthands and varied terminology, mapping disparate natural language phrases to the same underlying structured form~\cite{hsu2025llm,sadia2025squid}. 

Beyond explicit extraction, their pretraining also makes LLMs effective imputation engines for inferring latent events and implicit affordances that human writers typically omit for brevity. Moreover, LLMs also embed broad domain knowledge about these clinical narratives. They have been used to filter out implausible object interactions and state changes to prune the massive symbolic search spaces~\cite{valmeekam2023planning}. While LLMs excel at linguistic extraction and imputation, they fundamentally struggle with inductive logical reasoning~\cite{helff2025slr}, especially in complex domains involving interactions over long horizons~\cite{tantakoun2025llms}. As a result, they produce overly simplistic models.  Moreover, recent LLM-based planning formalizers typically assume at least one source of support unavailable when building models from clinical narratives, such as a clean domain description, executable environment feedback, or benchmark tasks with known formal structure~\cite{guan2023worldmodels,mahdavi2024environment,zhu2025psalm}, making them inapplicable to the task. Finally, most of these LLM+PDDL works have focused on deterministic domain generation~\cite{huang2025limit, oswald2024llmdomaingenerators,zhang2024proc2pddl},  whereas our setting requires a probabilistic PPDDL domain model to capture stochastic effects across clinical narratives. Unlike prior single-instance LLM formalizers, NSPIN uses LLMs for extraction and imputation from raw clinical narratives, while relying on symbolic induction and corpus-level aggregation to construct and validate probabilistic planning models.

\section{The NSPIN Framework} 
Recall our goal is to {\em induce a structured model of a clinical procedure from a database of unstructured notes}. We formally define the problem as follows:

\noindent\fbox{
    \begin{minipage}{0.93\linewidth}
\textbf{Given}:  A Database $\mathcal{D} = \{x^{(i)}\}_{i=1}^N$, where each $x^{(i)}$ is an unstructured clinical narrative describing a procedure instance.\\
\textbf{To Do}: A Probabilistic Planning Domain Model $\mathcal{P}$ that characterizes the actions, preconditions, and stochastic effects of the procedure.
\end{minipage}
}

This task presents two key challenges. First, clinical narratives consist of unstructured text, while inducing planning models requires structured action and observation data. Second, the narratives are highly heterogeneous and sparse, since clinicians have different documenting styles and underlying assumptions, which affect which steps are explicitly recorded and which are treated as implicit knowledge.

To address these challenges, we introduce NSPIN, a neurosymbolic framework for planning model induction. NSPIN bridges the representational gap by exploiting a pretrained LLM for extraction and symbolic refinement for logical grounding. Figure \ref{fig:neurosymbolic_pipeline} illustrates the structure of NSPIN, and the subsequent sections describe each stage in detail.

\subsection{Schema Creation} 
The schema generation stage introduces and defines the formal vocabulary used to represent the planning domain. In this step, an LLM takes as input a natural language description of the domain, including its rules and constraints, and generates a structured domain schema $\mathcal{P}_0$. This schema consists of three components: 
\\\textbf{1) Typed Object Categories (O):} A set of semantic classes (e.g., \texttt{location}, \texttt{structure}, \texttt{instrument})  that categorize the objects in the surgical environment.
\\\textbf{2) Observation Predicates ($\mathcal{F}$):} A finite set of first-order logic predicates that describe the properties of objects and their mutual relations, such as $\texttt{is\_occluded}(?structure)$. These predicates define the state space of the procedure.\\
\\\textbf{3) Action Predicates ($\mathcal{A}_0$):} 
A set of parameterized action types that name the available surgical steps, for example, $\texttt{mobilize\_structure}(?structure),$ where each argument is typed by an object category from $O$.

Note that the action schemas produced at this stage ($\mathcal{A}_0$) are initialized with empty preconditions and effects, $\text{pre}(a) = \text{eff}(a) = \emptyset$ for each $a \in \mathcal{A}_0.$ These are syntactically valid but vacuous placeholders, since each action is always applicable and has no effect on the state. This stage fixes only the vocabulary and argument structure of each action. The substantive preconditions and stochastic effects are induced from data in the planning model construction stage. This constrains the massive symbolic search space to a fixed, clinically relevant vocabulary by exploiting the pretrained LLM's approximate domain knowledge.


\subsection{Sequence Generation}
\label{sec:sequence_generation}

The sequence generation stage converts the unstructured surgical notes to structured observation and action traces required to induce the planning model. This process is divided into \textbf{information extraction} and \textbf{sequence imputation} phases.

\paragraph{Information Extraction.}
We formulate information extraction as a constrained grounding problem. Given an unstructured narrative $x \in \mathcal{D}$ and the schema $\mathcal{P}_0=\langle \mathcal{F}, O, \mathcal{A}_0 \rangle$, the LLM is prompted to construct a sequence of grounded predicates $s = [u_1, \dots, u_k]$
where each $u = h({\bf v})$ combines a predicate $h \in \mathcal{F}\cup\mathcal{A}_0$ with entities ${\bf v} = v_1,\dots,v_n$ grounded in $x,$ as per the schema. Since LLM outputs are stochastic, we generate K candidate solutions for the same note ${\bf s} = \{ s_1,  \dots s_K\}$ (lowercase, one per LLM sample),  and aggregate them to obtain a single consensus sequence, denoted as $S_0$ (uppercase), which we construct in three stages:
\\\textbf{1) Majority Filtering:} We retain only those predicate instances supported by a majority of the K candidates. Concretely, we retain
$${\bf h} = \{ (h, r):\  h \in  \mathcal{F}\cup\mathcal{A}_0,\ 1\leq r\leq r^*(h) \},$$  where $r^*(h)$ is the maximum number of instances of $h$ that are supported by more than half of the $K$ candidates, $$r^*(h) = \max \{  r: \text{Count}(h, r, {\bf s}) > \frac{K}{2},\ r \geq 0\}$$ and $\text{Count}(h, r, {\bf s})$ is the number of candidates $s \in {\bf s}$ containing at least $r$ occurrences of $h$. 
\\\textbf{2) Consensus Grounding:} For each retained predicate instance $(h, r) \in {\bf h},$ the value of argument slot $v_i$ is chosen by majority vote across the candidates,
$$v_i = \underset{v}{\arg\max}\  \text{Count}_i(v, h, r, {\bf s}),$$ where $\text{Count}_i(v, h, r, {\bf s})$ is the number of candidates $s \in {\bf s}$ where the $r$-th occurrence of predicate $h$ has $i$-th argument equal to $v.$ 
\\\textbf{3) Temporal Ordering:} To order the retained predicate instances $\mathbf{h}$, we solve a weighted Kemeny rank aggregation problem~\cite{kemeny_rank}. For $(h,r), (h',r') \in \mathbf{h}^2$, we define the pairwise precedence weight as,
$$w\big((h,r), (h',r')\big) = \text{Count}\big((h,r) \prec (h',r'),\ {\bf s}\big)$$
which counts the number of candidates $s \in \mathbf{s}$ in which the $r$-th occurrence of $h$ precedes the $r'$-th occurrence of $h'$. The aggregated sequence ${S}_0$ is the ordering $\sigma$ of $\mathbf{h}$ that minimizes the total contradicted precedence weight,
$${S}_0 = \arg\min_{\sigma} \sum_{(h,r) \prec_\sigma (h',r')} w\big((h',r'), (h,r)\big).$$


\paragraph{Sequence Imputation.} 
While the aggregated sequence $S_0$ captures the actions and observations mentioned in the surgical note, surgeons frequently omit obvious intermediary steps or universal surgical protocols. As a result, the extracted sequence $S_0$ lacks the logical continuity required for PPDDL induction. Some transitions between consecutive instances $(h,r), (h',r') \in S_0$ presuppose intermediate actions or observations that are not explicit in $x.$ Identifying these omitted steps requires domain knowledge of what would have been obvious to the surgeon and therefore not worth recording.

We address this by exploiting a pretrained LLM's approximately correct domain knowledge of clinical workflows to infer these omitted steps.
Given the incomplete sequence $S_0$, the LLM is prompted to infer additional grounded instances $u = h(\mathbf{v})$, $h \in \mathcal{F}\cup\mathcal{A}_0$, that are logically necessary or contextually implied by the surrounding instances in $S_0$, with reprompting in case of syntactic or schema violation. 

As in extraction, we generate $K$ candidate imputations for the same sequence $S_0$, yielding $K$ enriched candidate sequences $s_1', \dots, s_K'$, each containing the original instances of $S_0$ along with the proposed additions. To ensure robustness, we apply the same three-stage aggregation procedure used in extraction over $s_1',\dots,s_K'$ to obtain the final enriched sequence $S.$ Imputed instances are retained only if supported by a majority of candidates, their arguments are set by consensus vote, and their position relative to all other instances is fixed by Kemeny aggregation. Since every instance in $S_0$ is present in all $K$ enriched candidates by construction, it is trivially retained by majority filtering, so $S$ strictly extends $S_0$ with additional consensus-supported instances. This sequence represents the complete surgical workflow used for planning model construction.


\vspace{-0.5em}
\subsection{Planning Model Construction}
This stage transforms the grounded predicate sequences ${\bf S} = \{S^{(i)}\}_{i=1}^N$ into a planning domain model $\mathcal{P}$. It consists of two steps: a data-driven induction step that constructs an initial action schema from ${\bf S}$, followed by an LLM-based refinement step that revises the induced preconditions. The result is a planning model $\mathcal{P}$ represented in PPDDL\footnote{The algorithmic pseudocode for the different components of planning model construction is provided in the supplementary material at \url{https://github.com/s-ranveer/nspin_ppddl}}.

\paragraph{Data-Driven Induction.} Given the predicate sequences ${\bf S}$, we induce an initial PPDDL action schema $\alpha_a$ for each action $a \in \mathcal{A}_0$, specifying its preconditions and probabilistic effects from empirical co-occurrence and outcome frequencies. This induction step is modular, allowing any algorithm that learns action preconditions and effects from structured, action-segmented sequences to be used. We consider a greedy, coverage-based inducer~\cite{vasu2026localtogloballogicalexplanationsdeep} to construct preconditions from observation contexts of size $w$ till atleast a fraction $\tau_{obs}$ of sequences are satisfied; we estimate effects conditioned on the action arguments via maximum-likelihood outcome frequencies, filtering out any effects with a probability less than $\tau_{prob}.$ The resulting schemas are assembled into the domain $\mathcal{P}_1 = \langle \mathcal{F}, O, \mathcal{A}_1 \rangle$, where $\mathcal{A}_1 = \{\alpha_a \mid a \in \mathcal{A}_1\}$.

\subsubsection{LLM Refinement.}
While the data-driven action schema fits the observed data, the learned preconditions may be overly restrictive, including clauses that are merely associational, appearing before an action in the data without being genuinely necessary for that action to occur. We therefore refine each action's preconditions using an LLM to obtain constraints that are more semantically grounded in the narrative and less brittle.

Starting from the domain $\mathcal{P}_1$, we split the sequences $\mathbf{S}$ into training and validation sets and track the best domain found so far. For each action $a$ with failed precondition instances, we extract its satisfied and failed contexts, $C_a^+$ and $C_a^-$, from the training set, and prompt the LLM to propose a revised precondition. We accept a revised precondition only if it does not worsen precondition satisfaction or degrade validation performance, yielding the refined domain $\mathcal{P}=\langle \mathcal{F}, O, \mathcal{A} \rangle.$

\section{Empirical Evaluation}
We now present our empirical evaluations and aim to answer the following questions. While the first three questions focus on the algorithmic contributions, the last one focuses on the qualitative evaluation for the societal impact.
\\\textbf{Q1)} Does LLM-based imputation of implicit information yield more accurate domain models?
\\\textbf{Q2)} Does LLM-refinement help NSPIN induce better planning models?
\\\textbf{Q3)} Does NSPIN induce better planning domain models from clinical narratives than LLMs?
\\\textbf{Q4)} How well do NSPIN's imputed steps, induced preconditions, and generated narratives align with clinical knowledge and practice?

\begin{figure}
    \centering
    \includegraphics[width=0.95\linewidth]{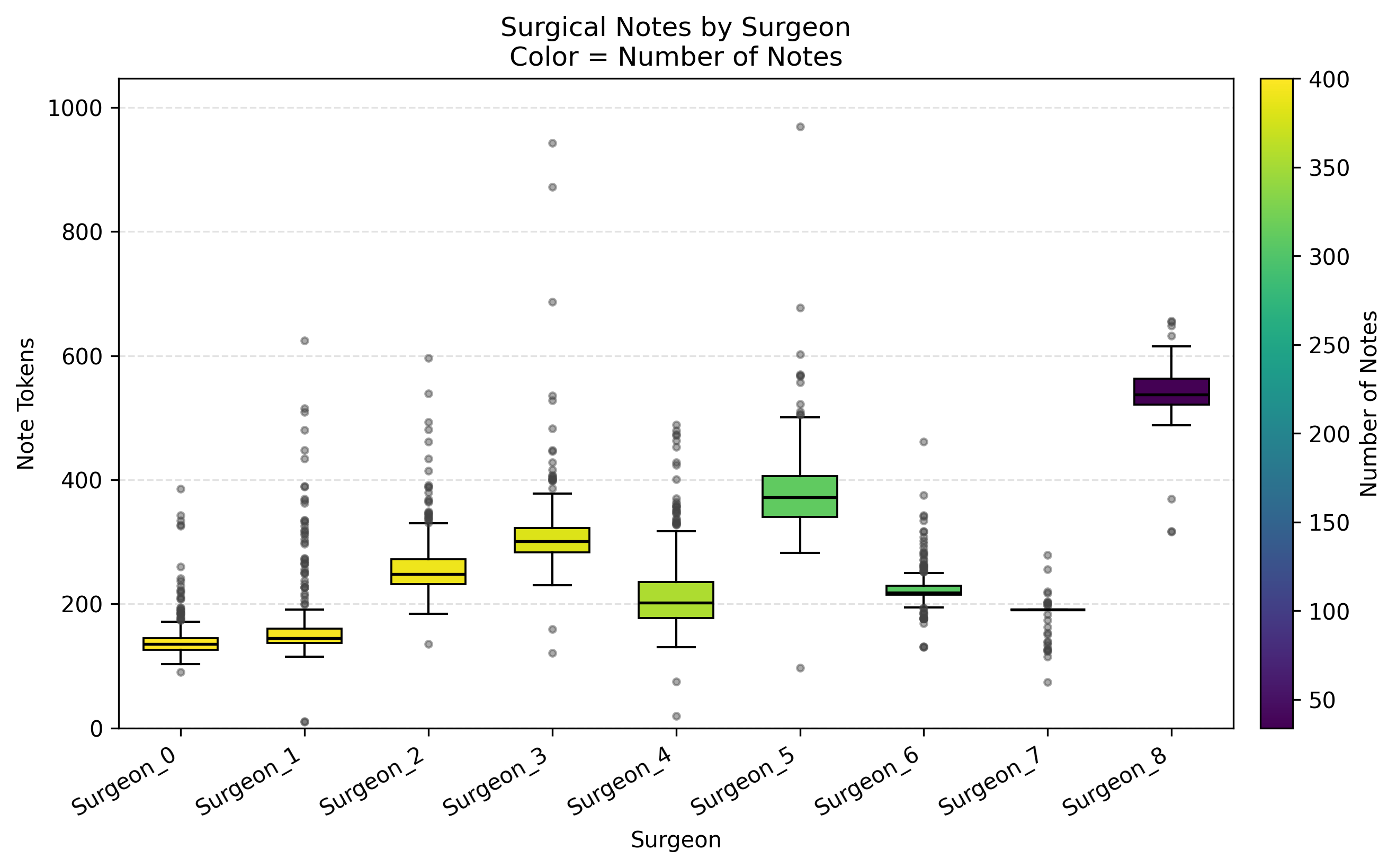}
    \caption{Surgical Note lengths and counts across surgeons}
    \label{fig:surgeon_distribution}
\end{figure}

\textbf{Dataset.} To answer these questions, we obtained a dataset of 2,660 laparoscopic appendectomy surgical notes from 9 different surgeons at Children's Hospital Medical Center in Dallas, TX\footnote{We are not aware of any publicly available equivalent datasets}. Each surgeon differs in the number of notes and average tokens per note, as shown in Figure \ref{fig:surgeon_distribution}.

\textbf{Methods.} 
We evaluate the probabilistic planning domains induced by NSPIN against two LLM-based baselines: {\bf LLM-only} and {\bf LLM-Example}. In LLM-only, the model is provided with the domain description and predicate templates and tasked with generating a PPDDL model zero-shot. In LLM-Example, the model is also provided with an example predicate sequence from the dataset to provide in-context grounding for the action schemas. Additionally, we also evaluate a variant of NSPIN without the final LLM-based precondition refinement.

To maintain the privacy of the sensitive clinical data, we employ a hybrid LLM strategy: Claude Sonnet 4.6~\cite{anthropic2025claudesonnet} is used for high-level schema construction and symbolic refinement, while the local, medically-tuned Medgemma-27b \cite{sellergren2026medgemmatechnicalreport} is used for the extraction and imputation of procedural traces from the raw narratives, with the extraction performed five times on each surgical note, and each extracted sequence independently imputed.

All LLM-based generations and refinement use Claude Sonnet. We use the Clingo Answer Set Programming (ASP) Solver to implement sequence merging\footnote{The different prompts, predicate templates, groundings, classification, code for running the different experiments, and the ASP program for the final sequence generation and the different PPDDLs created are provided in the supplementary material}of the five predicate sequences per patient based on the constraints defined before.

For data-driven PPDDL construction, we evaluated window sizes $(w \in {1, 3, 5, \infty}$), minimum effect-probability thresholds $(\tau_{\mathrm{prob}} \in {0.005, 0.01, 0.05})$, and minimum precondition-support thresholds $(\tau_{\mathrm{obs}} \in {0.7, 0.8, 0.9, 1.0})$, where $(\tau_{\mathrm{obs}})$ denotes the minimum fraction of sequences satisfying a candidate precondition. Grid search on the training set selected w = 3, $\tau_{\mathrm{prob}} = 0.01$, and $\tau_{\mathrm{obs}} = 0.8$.

\textbf{Metrics.}
We evaluate models along two dimensions. \textbf{Observation set prediction} measures how well a model's learned probabilistic effects predict the set of observations occurring between an action and the next, using a held-out trace. \textbf{Next-action prediction} measures how well a learned action-transition model predicts the subsequent action, while the domain's preconditions constrain which actions are symbolically applicable. For both, we report: \textbf{(1) Top-1/Top-3 accuracy}, whether the ground truth is the top prediction or among the top three; \textbf{(2) average negative log-likelihood (Avg NLL)}, the probability mass assigned to the ground truth, with lower values indicating better calibration; and, for next-action prediction, \textbf{(3) true action disallowed rate (TADR)}, the fraction of steps where the learned preconditions incorrectly rule out the ground-truth action, with lower values indicating better compatibility with observed workflows. We compute these metrics for each method using 9-fold cross-validation, where the PPDDLs are learnt from predicate sequences of all but one surgeon, and evaluated on the held-out surgeon's predicate sequences.

\subsection{Results} The generated appendectomy schema comprises 63 predicates in total, including 34 observation predicates, 26 action predicates, and 3 relational predicates, along with 58 typed argument categories and 322 enumerated grounding values. Tables \ref{tab:imputation}, \ref{tab:obs_results}, and \ref{tab:action_results} show the average evaluation results across 9 folds on the held-out set, and will be used for answering the questions asked at the beginning of the section. Statistical significance was assessed using the Wilcoxon signed-rank test with a significance level of $\alpha = 0.05$; all reported comparisons were statistically significant.
\input{Tables/impute_results}
\input{Tables/obs_results}
\input{Tables/action_results}

\textbf{(Q1) LLM-based Imputation of Sequences.}
To evaluate the utility of using LLMs for resolving sparsity in clinical narratives, we compare the action set prediction accuracy of data-driven models (NSPIN w/o refinement) induced from raw extractions versus those enriched via latent sequence imputation. As shown in Table \ref{tab:imputation}, we find that imputation improves action set predictions across all metrics. The model learned from imputed data had a higher Top-1 accuracy, Top-3 accuracy, and a lower average negative log likelihood. This consistent improvement indicates that the imputed sequences recover information that surgeons omitted as clinically obvious, and that this recovered information is essential.
Therefore, we can answer Q1 affirmatively.

\textbf{(Q2) LLM-Refinement in NSPIN.}
To assess whether LLM-based refinement of action preconditions can resolve the spurious correlation problem inherent in purely data-driven induction, we compare the predictive performance of NSPIN with a version without the LLM-based refinement. As shown in Table \ref{tab:obs_results}, observation-set prediction remained stable across both settings, as these metrics are conditioned on action execution and are independent of precondition logic. However, the refinement phase yielded substantial improvements in all next-action prediction metrics (Table \ref{tab:action_results}). Notably, the refined model achieved higher Top-1 and Top-3 accuracy and a lower Avg NLL, coupled with a significant reduction in the True Action Disallowed Rate (TADR). This indicates that the refinement step produced less restrictive preconditions that were more compatible with held-out traces. For example, for the step of sending the appendix specimen to pathology, the symbolic precondition inducer included descriptive findings such as associated pathology. The LLM-based refinement removed these while retaining the core procedural requirement that the specimen must first be retrieved, yielding a precondition that is more semantically faithful to the workflow. Overall, this improved next-action prediction and reduced TADR. So, we can answer Q2 affirmatively.

\textbf{(Q3) NSPIN vs LLM-Induced Models. } To evaluate the necessity of neurosymbolic integration, we compare the predictive validity of NSPIN against the purely LLM-based baselines (\textit{LLM-Only} and \textit{LLM-Example}). As shown in  Table \ref{tab:obs_results} and Table \ref{tab:action_results}, NSPIN outperforms the baselines in all metrics, achieving a higher Top-1 and Top-3 accuracy, as well as a lower average negative log likelihood. This indicates that the data-driven symbolic model provides a stronger explanation of post-action observations than domains generated directly by the LLM baselines.

The evaluation of next-action prediction reveals a more complex trajectory (Table \ref{tab:action_results}). The unrefined model (\textit{NSPIN w/o Refinement}) underperforms the neural baselines, exhibiting a higher TADR. However, following the one-step semantic refinement, NSPIN performance improves substantially, ultimately surpassing both neural baselines in predictive accuracy, likelihood, and TADR. This shift demonstrates that while LLMs possess strong procedural priors, they require symbolic validation to ground those priors in empirical data. Thus, we can answer Q3 affirmatively.


\textbf{(Q4) Qualitative evaluation.} To assess whether LLM-based imputation recovers clinically real but undocumented or missing steps, we considered the top-10  frequently imputed predicates and sampled 5 notes where they were imputed. Our domain expert found the imputation clinically correct. For example,  one commonly imputed predicate is $\texttt{fascia\_closure}$. In one of the notes, the information extractor failed to extract this despite it being made explicit by the sentence \textit{"The umbilical fascia was closed with 0 Vicryl suture."}. The imputer correctly recovered this information based on its clinical necessity and the partially extracted sequence. In another example, the predicate $\texttt{appendix\_base\_secured}$ was imputed by the LLM before transection ($\texttt{appendix\_transected}$). Although the note does not explicitly mention securing the appendix, it is typically performed with transection as part of the same mechanical action. Because transection cannot be completed without securing the appendix, the imputation is justified.

To assess whether the induced preconditions and their refinement reflect genuine clinical necessity, we selected three actions for expert review: \texttt{port\_placement,\  appendix\_base\_secured,} and \texttt{appendix\_removed}. 
Port placement requires that an incision already be made at the relevant site. The LLM-only baseline omitted the precondition completely, allowing ports to be placed anywhere as long as one has not already been placed. NSPIN, without refinement, listed it as a contributing but non-required condition. It was made the necessary condition only after refinement. 
Securing the appendix base requires the appendix structure to be mobilized. 
This precondition was missing even after refinement; mobilization remained a contributing rather than necessary condition in NSPIN both before and after refinement, while the requirement was completely absent for the LLM baselines. Finally, for appendix removal, the LLM-only baselines included stump treatment as a precondition. While stump treatment typically occurs before removal, it is not a necessary precondition as removal does not logically depend on it. NSPIN, with and without refinement, correctly identified the full set of relevant preceding actions, but some were rendered only as contributing rather than necessary actions due to the presence of certain observations acting as contributing preconditions as well.

Our expert found the effects of all 3 actions clinically valid. In contrast, the LLM-based baselines had unrealistic effect probabilities. For example, the probability of injury for port placement was substantially higher than what our expert considered realistic.    


To qualitatively test whether the induced PPDDL model captures sufficient clinical knowledge to generate coherent procedure descriptions, we had the expert review five representative surgical notes synthesized using the model. We sampled action sequences from the transition model learned during evaluation, using the PPDDL preconditions as guards to determine which actions could occur and emitting the corresponding observations for each sampled action, such as the one in Figure~\ref {fig:nspin_sampled_sequence}. An LLM with the predicate templates, the sampled sequence, and generation rules to generate synthetic notes. The generated notes generally preserved the high-level temporal structure of an appendectomy but sometimes missed some nuances that were underrepresented in the training data. For example, the notes would indicate making a suprapubic incision before placing a camera port, when a different incision would be warranted. These errors suggest that while the learned model captures the broad procedural workflow, it could benefit from targeted expert knowledge about infrequently observed constraints.

Taken together, these evaluations indicate that NSPIN recovers clinically real implicit steps, its refined preconditions align with clinical domain knowledge, and its symbolic structure can be used to synthesize plausible surgical narratives.
\section{Conclusion}
We considered the challenging problem of translating unstructured clinical narratives into a formal representation. We address two key challenges in the clinical planning domain induction: the linguistic variability inherent in natural language procedure notes and the observational sparsity caused by human reporting bias. Our work bridges the gap between incomplete, unstructured narratives and the structured representations required for clinical decision support by exploiting the broad priors encoded in pretrained LLMs. Our empirical evaluation on a corpus of 2,660 surgical notes on laparoscopic appendectomies written by 9 surgeons shows that NSPIN effectively addresses linguistic variability and reporting bias, yielding domain descriptions that generalize to unseen data -- a key requirement for many clinical tasks.

LLM-based sequence imputation improves downstream predictive performance in a manner consistent with recovering omitted or unmentioned procedural steps that clinicians frequently omit but are logically essential for plan consistency. Further, we showed that a single step of neural refinement significantly improves the generalizability of domain descriptions by pruning spurious temporal correlations that a purely data-driven approach would otherwise treat as hard preconditions. These results suggest that combining LLM-based extraction and imputation with symbolic induction is a promising approach for constructing useful PPDDL-like procedural models from clinical narratives.

There are several directions for future work. First, evaluation on additional surgical procedures and clinical domains beyond appendectomy is necessary. Second, the modular design of NSPIN makes it possible to explore alternative methods for inducing PPDDL models from structured predicate sequences. Finally, an important extension is to move beyond refining only action preconditions and develop methods for refining probabilistic effects as well. Overall, together with these directions, NSPIN represents a significant step toward intelligent clinical decision support systems capable of reasoning about complex medical procedures.

\section*{Acknowledgment}
The authors gratefully acknowledge support from the NIH award R01NS133142 and the Cluster of Excellence ``Reasonable AI" funded by the German Research Foundation (DFG) under Germany’s Excellence Strategy, EXC-3057.

\bibliography{bibliography}

\end{document}

%% file: Figures/framework_1.tex
\definecolor{llmblue}{RGB}{230, 240, 255}
\definecolor{logicred}{RGB}{255, 235, 235}
\definecolor{outputgreen}{RGB}{240, 255, 240}
\definecolor{grayborder}{RGB}{180, 180, 180}
\definecolor{arrowgray}{RGB}{80, 80, 80}

\begin{figure*}[htbp]
    \centering
    \resizebox{0.9\textwidth}{!}{
    \begin{tikzpicture}[
        transform shape,
        node distance=1cm,
        >=Stealth,
        pane/.style={
            draw=grayborder, 
            thick, 
            rectangle, 
            rounded corners=3pt, 
            fill=white, 
            text width=4.2cm, 
            minimum height=5.5cm,
            inner sep=10pt, 
            font=\small,
            align=left
        },
        procblock/.style={
            draw, 
            rectangle, 
            rounded corners=5pt, 
            minimum width=3.4cm, 
            minimum height=1.2cm, 
            align=center, 
            font=\small\bfseries,
            inner sep=5pt
        },
        mainarrow/.style={
            ->, 
            line width=2.5pt, 
            color=arrowgray!30
        },
        innerarrow/.style={
            ->, 
            thick, 
            color=arrowgray
        }
    ]

        \node[pane] (input) {
            {\large\bfseries Surgical Notes} \\
            \rule{\linewidth}{0.6pt} \\[8pt]
            \textit{``Following anesthesia, the patient was prepped and draped off sterilely. A supraumbilical incision was made.  A 12 mm port was placed ... ''}
        };
        \node[procblock, fill=llmblue, right=1.5cm of input.east, anchor=west, yshift=0.4375cm] (extraction) {
            Schema Creation
        };
        \node[below=4pt of extraction, text width=3.4cm, align=center, ] (ext_desc) {
            Creates the Domain Schema
        };
        
        \node[procblock, fill=llmblue!60, right=1.2cm of extraction.east, anchor=west] (sequence) {
            Sequence Generation 
        };
        \node[below=4pt of sequence, text width=3.4cm, align=center, ] (seq_desc) {
            Extracts predicate sequences from data\\ events temporally
        };
        
        \node[procblock, fill=logicred, right=1.2cm of sequence.east, anchor=west] (induction) {
            Planning Model \\ Construction
        };
        \node[below=4pt of induction, text width=3.4cm, align=center, ] (ind_desc) {
            Synthesizes formal \\ PDDL operators
        };
        
        \draw[innerarrow] (extraction) -- (sequence);
        \draw[innerarrow] (sequence) -- (induction);
        
        \begin{scope}[on background layer]
            \node[
                draw=gray!40, 
                dashed, 
                thick, 
                fill=gray!3, 
                rounded corners=10pt, 
                inner sep=15pt, 
                fit=(extraction) (sequence) (induction) (ext_desc) (seq_desc) (ind_desc)
            ] (bridge) {};
            \node[anchor=south, font=\large\bfseries, yshift=8pt] at (bridge.north) {NSPIN};
        \end{scope}

        \node[pane, right=.75cm of bridge.east, anchor=west, fill=outputgreen!40] (output) {
            {\large\bfseries Planning Model} \\
            \rule{\linewidth}{0.6pt} \\[8pt]
            \texttt{(:action port\_placement} \\
            \texttt{\ \ :parameters (?port\_size ?location)} \\
            \texttt{\ \ :precondition (and (done\_incision\_made))} \\
            \texttt{\ \ :effect (probabilistic} \\
            \texttt{\ \ \ \ 1 (and (done\_port\_placement))} \\
            \texttt{)}};
        \draw[mainarrow] (input) -- (bridge);
        \draw[mainarrow] (bridge.east) -- (output.west);

    \end{tikzpicture}}
    \caption{{\bf The NSPIN framework for inducing probabilistic planning models from clinical narratives.} Our neurosymbolic approach extracts structured predicates from unstructured surgical notes using an LLM. It uses this structured information to synthesize a planning model that generalizes across diverse surgical workflows. Using an LLM to ground each action and observation in the text allows NSPIN to account for both explicit and implicitly mentioned information.}
    \label{fig:neurosymbolic_pipeline}
\end{figure*}

%% file: Tables/impute_results.tex
\begin{table}[!t]
\centering
\begin{tabular}{llll}
\hline
\textbf{Dataset} & \textbf{Top-1 Acc} & \textbf{Top-3 Acc} & \textbf{Avg NLL} \\
& \textbf{$\uparrow$ is better} & \textbf{$\uparrow$ is better} & \textbf{$\downarrow$ is better} \\
\hline
Imputed & $\mathbf{0.32\scriptstyle{\pm0.07}}$ & $\mathbf{0.51\scriptstyle{\pm0.10}}$ & $\mathbf{9.33 \scriptstyle{\pm 2.35}}$ \\
Non-Imputed & $0.24\scriptstyle{\pm 0.09}$ & $0.39\scriptstyle{\pm 0.12}$ & $11.85\scriptstyle{\pm 2.85}$ \\
\hline
\end{tabular}
\caption{Action set prediction performance comparison across imputed and non-imputed datasets}
\label{tab:imputation}
\end{table}

%% file: Tables/obs_results.tex
\begin{table*}[!t]
\centering
\begin{tabular}{l l l l}
\hline
\textbf{Method} & \textbf{Top-1 Acc} & \textbf{Top-3 Acc} & \textbf{Avg NLL} \\
 & \textbf{$\uparrow$ is better} & \textbf{$\uparrow$ is better} & \textbf{$\downarrow$ is better} \\
\hline

NSPIN (w/wo) Refinement
&$\mathbf{0.88 \scriptstyle{\pm 0.03}}$ & $\mathbf{0.93 \scriptstyle{\pm 0.03}}$ & $\mathbf{1.67 \scriptstyle{\pm 0.60}}$ \\

LLM with Example & $0.56 \scriptstyle{\pm 0.23}$ & $0.56 \scriptstyle{\pm 0.23}$ &  $9.22 \scriptstyle{\pm 4.76}$ \\

LLM only
& $0.78 \scriptstyle{\pm 0.09}$ & $0.78 \scriptstyle{\pm 0.09}$ & $4.72 \scriptstyle{\pm 1.80}$ \\\hline
\end{tabular}
\caption{Observation prediction performance comparison on the imputed dataset across different methods}
\label{tab:obs_results}
\end{table*}

%% file: Tables/action_results.tex
\begin{table*}[!t]
\centering
\begin{tabular}{lllll}
\hline
\textbf{Method} & \textbf{Top-1 Acc} & \textbf{Top-3 Acc} & \textbf{Avg NLL} & \textbf{TADR} \\
& \textbf{$\uparrow$ is better} & \textbf{$\uparrow$ is better} & \textbf{$\downarrow$ is better} & \textbf{$\downarrow$ is better} \\
\hline
NSPIN  
& $\mathbf{0.42 \scriptstyle{\pm 0.07}}$ & $\mathbf{0.64 \scriptstyle{\pm 0.08}}$ & $\mathbf{6.08 \scriptstyle{\pm 1.57}}$ & $\mathbf{0.16 \scriptstyle{\pm 0.06}}$ \\
NSPIN  w/o Refinement
& $0.32 \scriptstyle{\pm 0.07}$ & $0.51 \scriptstyle{\pm 0.09}$ & $9.33 \scriptstyle{\pm 2.22}$ & $0.36 \scriptstyle{\pm 0.12}$ \\
LLM with Example & $0.09 \scriptstyle{\pm 0.16}$ & $0.19 \scriptstyle{\pm 0.27}$ & $16.28 \scriptstyle{\pm 6.34}$ & $ 0.74 \scriptstyle{\pm 0.36}$ \\
LLM only & $0.20\scriptstyle{\pm 0.13}$ & $0.27 \scriptstyle{\pm 0.16}$ & $15.05 \scriptstyle{\pm 3.49}$ & $ 0.68 \scriptstyle{\pm 0.19}$ \\
\hline
\end{tabular}
\caption{Action prediction performance comparison on the imputed dataset across different methods} \vspace{-0.5em}
\label{tab:action_results}
\end{table*}